\def\year{2022}\relax
\documentclass[letterpaper]{article} 
\usepackage{aaai22}  
\usepackage{times}  
\usepackage{helvet}  
\usepackage{courier}  
\usepackage[hyphens]{url}  
\usepackage{graphicx} 
\usepackage{natbib}  
\usepackage{caption} 
\DeclareCaptionStyle{ruled}{labelfont=normalfont,labelsep=colon,strut=off} 
\usepackage{algorithm}
\usepackage{algpseudocode}

\usepackage{newfloat}
\usepackage{listings}
\floatstyle{ruled}
\newfloat{listing}{tb}{lst}{}
\floatname{listing}{Listing}
\usepackage{subcaption}
\usepackage{tikz} 
\usepackage{amsmath}
\DeclareMathOperator*{\argmax}{argmax} 

\usepackage[retainorgcmds]{IEEEtrantools}

\usepackage{booktabs}
\usepackage{multirow}

\usepackage[figuresleft]{rotating}
\usepackage{paralist}

\usepackage{mathtools}
\usepackage{amssymb}

\title{Personalized Public Policy Analysis in Social Sciences\\Using Causal-Graphical Normalizing Flows}
\author {
    Sourabh Balgi\textsuperscript{\rm 1}, 
    Jose M. Pe{\~n}a\textsuperscript{\rm 1}, 
    Adel Daoud\textsuperscript{\rm 2,}\textsuperscript{\rm 3} \\
}
\affiliations {
    \textsuperscript{\rm 1} The Department of Computer and Information Science (IDA), Link{\"o}ping University, Sweden \\
    \textsuperscript{\rm 2} The Institute for Analytical Sociology (IAS), Link{\"o}ping University, Sweden \\
    \textsuperscript{\rm 3} The 
Department of Computer Science and Engineering,
Chalmers University of Technology, Gothenburg, Sweden \\
    sourabh.balgi@liu.se, jose.m.pena@liu.se, adel.daoud@liu.se
}

\usepackage{bibentry}

\begin{document}

\maketitle

\begin{abstract}

Structural Equation/Causal Models (SEMs/SCMs) are widely used in epidemiology and social sciences to identify and analyze the average causal effect (ACE) and conditional ACE (CACE). Traditional causal effect estimation methods such as Inverse Probability Weighting (IPW) and more recently Regression-With-Residuals (RWR) are widely used - as they avoid the challenging task of identifying the SCM parameters - to estimate ACE and CACE. However, much work remains before traditional estimation methods can be used for counterfactual inference, and for the benefit of Personalized Public Policy Analysis (P3A) in the social sciences. While doctors rely on personalized medicine to tailor treatments to patients in laboratory settings (relatively closed systems), P3A draws inspiration from such tailoring but adapts it for open social systems. In this article, we develop a method for counterfactual inference that we name causal-Graphical Normalizing Flow (c-GNF), facilitating P3A. A major advantage of c-GNF is that it suits the open system in which P3A is conducted. First, we show how c-GNF captures the underlying SCM without making any assumption about functional forms. This capturing capability is enabled by the deep neural networks that model the underlying SCM via observational data likelihood maximization using gradient descent. Second, we propose a novel dequantization trick to deal with discrete variables, which is a limitation of normalizing flows in general. Third, we demonstrate in experiments that c-GNF performs on-par with IPW and RWR in terms of bias and variance for estimating the ACE, when the true functional forms are known, and better when they are unknown. Fourth and most importantly, we conduct counterfactual inference with c-GNFs, demonstrating promising empirical performance. Because IPW and RWR, like other traditional methods, lack the capability of counterfactual inference, c-GNFs will likely play a major role in tailoring personalized  treatment, facilitating P3A, optimizing social interventions - in contrast to the current `one-size-fits-all' approach of existing methods.
\end{abstract}

\section{Introduction}\label{sec:intro}
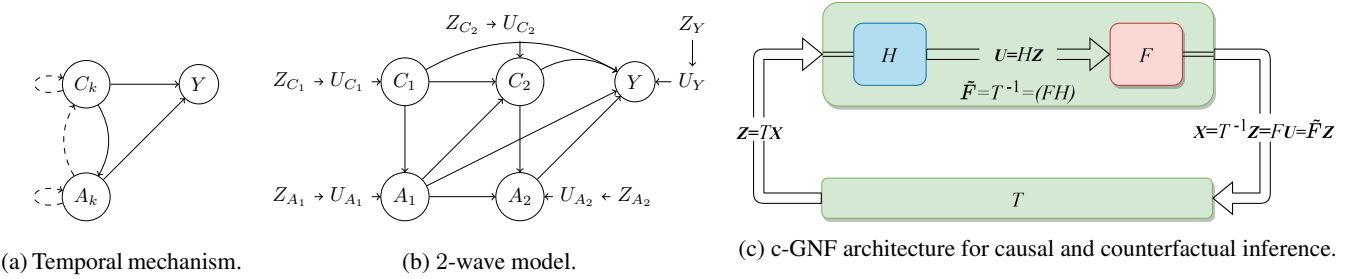
\begin{figure*}[ht!]
\begin{subfigure}[t]{0.18\textwidth}
{ 
\begin{center}
\resizebox{0.85\textwidth}{!}{
\begin{tikzpicture}
\tikzset{vertex/.style = {shape=circle,draw,minimum size=1.5em}}
\tikzset{edge/.style = {->}}
\node[vertex] (C_k) at (0,-2) {$C_k$};
\node[vertex] (A_k) at (0,-4) {$A_k$};
\node[vertex] (Y) at (2,-2) {$Y$};
\draw[edge,bend left,dashed] (A_k) to (C_k);
\draw[edge,bend left] (C_k) to (A_k);
\draw[edge] (C_k) to (Y);
\draw[edge] (A_k) to (Y);
\draw[edge,loop left,dashed] (C_k) to (C_k);
\draw[edge,loop left,dashed] (A_k) to (A_k);
\end{tikzpicture} 
}
\end{center}
\caption{Temporal mechanism.}
\label{sfig:nh_mechanism}
}
\end{subfigure}
\hfill
\begin{subfigure}[t]{0.36\textwidth}
{ 
\begin{center}
\resizebox{0.95\textwidth}{!}{
\begin{tikzpicture}
\tikzset{vertex/.style = {shape=circle,draw,minimum size=1.5em}}
\tikzset{edge/.style = {->}}
\node[vertex] (C_1) at (0,-2) {$C_1$};
\node[vertex] (C_2) at (2,-2) {$C_2$};
\node[vertex] (A_1) at (0,-4) {$A_1$};
\node[vertex] (A_2) at (2,-4) {$A_2$};
\node[vertex] (Y) at (4,-2) {$Y$};
\node[] (U_C_1) at (-1,-2) {$U_{C_1}$};
\node[] (U_C_2) at (2,-1) {$U_{C_2}$};
\node[] (U_A_1) at (-1,-4) {$U_{A_1}$};
\node[] (U_A_2) at (3,-4) {$U_{A_2}$};
\node[] (U_Y) at (5,-2) {$U_{Y}$};
\node[] (Z_C_1) at (-2,-2) {$Z_{C_1}$};
\node[] (Z_C_2) at (1,-1) {$Z_{C_2}$};
\node[] (Z_A_1) at (-2,-4) {$Z_{A_1}$};
\node[] (Z_A_2) at (4,-4) {$Z_{A_2}$};
\node[] (Z_Y) at (5,-1) {$Z_{Y}$};
\draw[edge] (Z_C_1) to (U_C_1);%
\draw[edge] (Z_C_2) to (U_C_2);%
\draw[edge] (Z_A_1) to (U_A_1);%
\draw[edge] (Z_A_2) to (U_A_2);%
\draw[edge] (Z_Y) to (U_Y);%
\draw[edge] (U_C_1) to (C_1);%
\draw[edge] (U_C_2) to (C_2);%
\draw[edge] (U_A_1) to (A_1);%
\draw[edge] (U_A_2) to (A_2);%
\draw[edge] (U_Y) to (Y);%
\draw[edge, bend left] (C_1) to (Y);
\draw[edge, bend left] (C_2) to (Y);
\draw[edge] (C_1) to (A_1);
\draw[edge] (C_2) to (A_2);
\draw[edge] (A_1) to (Y);
\draw[edge] (A_2) to (Y);
\draw[edge] (A_1) to (C_2);
\draw[edge] (C_1) to (C_2);
\draw[edge] (A_1) to (A_2);
\end{tikzpicture} 
}
\end{center}
}
\caption{$2$-wave model.}
\label{sfig:nh_mechanism 2-wave}
\end{subfigure}
\hfill
\begin{subfigure}[t]{0.45\textwidth}
{ 
\begin{center}
\includegraphics[width=\linewidth,keepaspectratio]{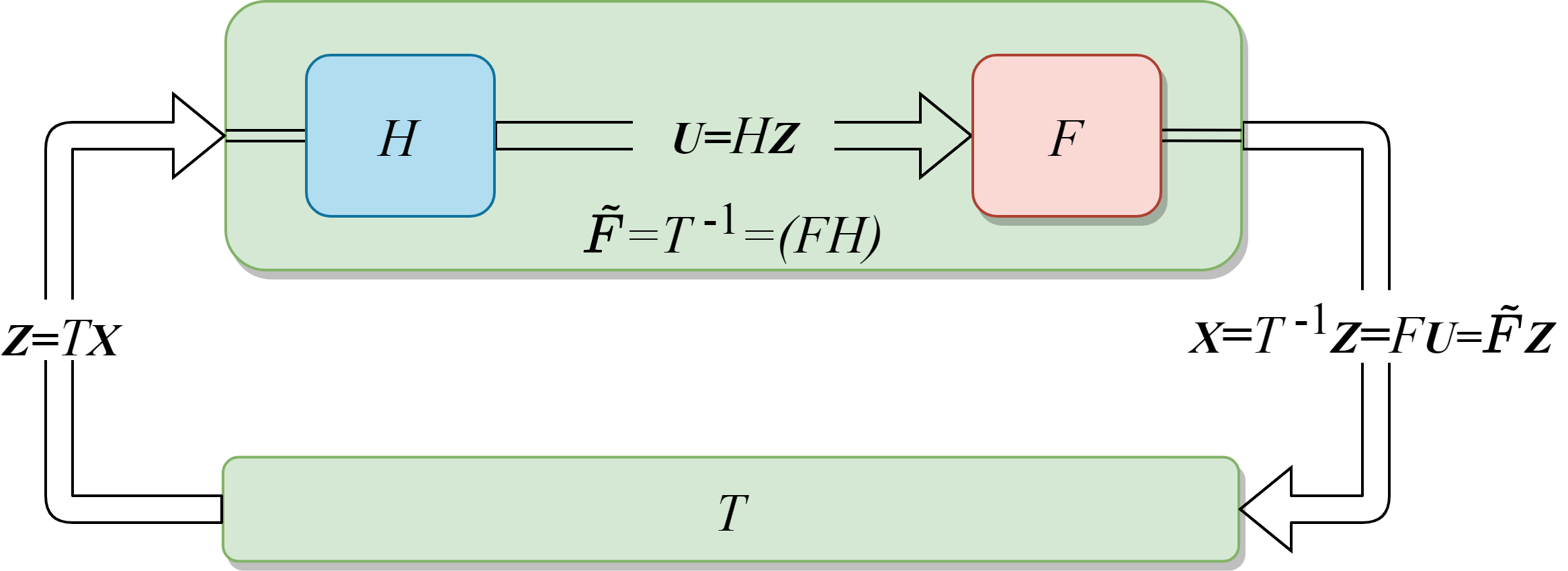}
\caption{c-GNF architecture for causal and counterfactual inference.}
\label{sfig:c-GNF arch}
\end{center}
}
\end{subfigure}
\caption{Fig.~\ref{sfig:nh_mechanism} shows temporal interactions between the treatments $A_k$ and the observed covariates/confounders $C_k$ at time-wave $k{ = }1,\ldots,K$  and the outcome of interest $Y$. The solid edges indicate the cause and effect at the same time-wave and the dashed edges indicate the one time-wave delayed effect with respect to the respective causes. 
Fig.~\ref{sfig:nh_mechanism 2-wave} represents the 2-wave model of Fig.~\ref{sfig:nh_mechanism}.
For any given observed endogenous variable $X$ in Fig.~\ref{sfig:nh_mechanism 2-wave}, $U_X$ and $Z_X$ respectively denote the unobserved exogenous noises of the true SCM $F{ : }\mathbf{U}{ \rightarrow }\mathbf{X}$ and the \emph{encapsulated-SCM} $\tilde{F}{ = }(FH){ = }T^{-1}{ : }\mathbf{Z}{ \rightarrow }\mathbf{X}$ in Fig.~\ref{sfig:c-GNF arch}, where $H{ : }\mathbf{Z}{ \rightarrow }\mathbf{U}$ denotes an auxiliary transformation. Since $\tilde{F}$ (green) encapsulates $F$ (red) and $H$ (blue) SCMs, i.e., $\tilde{F}{ = }(FH){ = }T^{-1}$, we refer to $\tilde{F}$ as the \emph{encapsulated-SCM}.
Our c-GNF models $T{ : }\mathbf{X}{ \rightarrow }\mathbf{Z}$ and readily provides the \emph{encapsulated-SCM} $\tilde{F}{ : }\mathbf{Z}{ \rightarrow }\mathbf{X}$, as $T$ is invertible by construction, facilitating counterfactual inference using \emph{`The First Law of Causal Inference'}~\cite{pearl2009abductionactionprediction, pearl2009causality}.
}
\label{fig:nh_causaC_model all}
\end{figure*}

Since the realization that correlation does not imply causation~\cite{wright1921correlationcausation, fisher1936designofexperiments}, statisticians, computer scientists, and social scientists have been developing ways to identify and estimate causal effects from observational data.
By the 1970s, Structural Equation Models (SEMs), developed from Wright's path analysis~\cite{wright1921correlationcausation} in genetics, have been widely used in economics~\cite{Haavelmo1943SEM, Goldberger1972SEMecon} and other social sciences~\cite{Duncan1974SEMsocio, Ploch1975SEM, Fienberg1975Introduction2SEM} to analyze cause and effect. 
Pearl's Structural Causal Model (SCM)~\cite{pearl2009causality} provides an explicit causal definition to disambiguate the causes and effects by replacing the symmetric `=' operator with asymmetric `:=' assignment operator, thereby complementing the original SEM definition.
Energized by this and similar developments in computer science, a causal revolution has occurred with large ramifications not only in academia but also for how governments, non-governmental organizations, and international organizations (such as the World Bank, United Nations Children’s Fund (UNICEF), or International Monetary Fund) articulate public policies to combat social ills. For example, to combat poverty, governments use mainly an individual's income and similar characteristics to identify vulnerable population and then assess if they are eligible for social welfare eligibility. Then a simple rule is often applied: if they are eligible, they tend to receive a fixed one-size-fits-all public policy; if they are not eligible, no policy is ascribed. Although such public-policy making is highly transparent as it applies what is best on average for a population -- that is the average causal effect (ACE) -- it lacks an adaptability necessary to efficiently combat poverty, ill-health, and other social ills. For effective combating, government officials and others require methods that are able to personalize these policies~\cite{kino2021adelmlreview}. That is, personalized public policy analysis (P$^3$A) requires methods that can move beyond ACE estimation, and into counterfactual inference.

However, traditional statistical causal effect estimation methods, such as Inverse Probability Weighting (IPW)~\cite{Rosenbaum1983propensityscoreipw, hernan2009ipw} or recent ones such as Regression-With-Residuals (RWR)~\cite{wodtke2020rwr}, focus on ACE estimation only. An advantage of IPW (stabilized) is that it provides a simple and effective way to estimate ACE without the need to model the entire causal system, under the assumption that the functional form of the propensity score is correctly specified. In contrast to IPW, RWR, which comes under the class of outcome regression models, estimate ACEs more effectively (less variance), under the assumption that the functional form of the outcome models are correctly specified. 
However due to the use of a single outcome model similar to S(Single)-Learner, RWR provides biased estimates when there is effect modification in data (also know as effect heterogeneity)~\cite{vanderweele2009distinctionheterogeneity}.  

To directly address effect modification, computer scientists have developed a class of outcome regression based machine-learning models called meta-learners~\cite{kunzel2019metalearners} (e.g., T(Two)-Learner, X-Learner) where each treatment group is modeled using a separate outcome regression models conditioned on proper adjustment sets to account for any confounding.
Since meta-learners model separate conditional outcome regression models for each treatment groups, we alternatively refer them as Grouped Conditional Outcome Model (GCOM).
The conditioning set or adjustment set or the predictors of outcome regression model are identified from the non-parametric expression of the interventional distribution obtained from $do$-calculus~\cite{pearl2009causality, pearl2012docalculus} or $G$-computation formula~\cite{ROBINS1986gcom, hernan2009ipw} to accurately model and estimate the causal effects of interest by adjusting for the confounding.
While IPW requires the right propensity score model, RWR and GCOM methods require the correct outcome model. 
In contrast, Doubly Robust (DR)~\cite{bang2005doublyrobust} methods are a class of causal inference methods that incorporate the best of IPW and GCOM by jointly modeling the propensity score and the outcome. The key advantage of DR is that it requires only either one of the propensity or the outcome model to be  correctly specified to produce unbiased ACE estimates. Notwithstanding the advantages of IPW, RWR, GCOM, and DR, none of them successfully supply a method to conduct counterfactual inference for P$^3$A. 

In this work, we aim to further P$^3$A and causal inference in the social science by demonstrating a method for counterfactual inference that does not require a priori knowledge of the correct functional form in neither the propensity score or the outcome model. Our causal method consists of a flow-based deep learning approach known as Graphical Normalizing Flows (GNFs). As we re-purpose GNFs for causal inference by using causal-Directed Acyclic Graphs (causal-DAGs), we refer them as causal-GNFs (c-GNFs). Even though we focus on the social effect of neighborhood on children's education outcomes~\cite{wodtke2020rwr} as an example of P$^3$A in this article, our contributions are applicable to other settings in epidemiology, economics, sociology, and political science, where counterfactual policy-making is critical to promote social, material, and health outcomes. 
Our work supplies at least the following five contributions.

\begin{inparaenum}[1.]
    \item We show that c-GNF models the underlying SCM of interest in the form of an \emph{encapsulated-SCM} using only the observational data and without making assumptions on the unobserved exogenous noises or the functional mechanisms of the true SCM.
    
    \item Unlike IPW and RWR that require correct functional forms to provide unbiased ACEs, we show that c-GNF needs no functional form assumptions to achieve unbiasedness because the encapsulated-SCM, parameterized by a deep neural networks, has the ability to model any functional form from the observational data via likelihood maximization by gradient descent.
    
    
    \item We demonstrate experimentally that c-GNF is on-par with IPW and RWR in terms of the bias and variance of the ACE estimates when the right functional forms are assumed and above-par when they are unknown under both small and large sample sizes.
    
    \item Normalizing flows (NFs) are generally built for continuous variables. As this challenges the use of discrete variables, we propose a novel yet simple \emph{Gaussian dequantization} trick to adapt discrete variables for NFs.

    \item The most important, unique and key contribution of our c-GNF is that it has the capacity to perform counterfactual inference since c-GNF models the encapsulation of the true SCM. This enables scholars and policymakers to move beyond merely identifying the optimal treatment (public policy) at the population level and start tailoring individual-level optimal treatment (public policy). This capacity will advance P$^3$A in social science and other domains (such as personalized medicine)~\cite{kino2021adelmlreview}.
    
\end{inparaenum}

In the next section, we introduce the notations, define a typical social problem, and discuss the assumptions required to solve it. Then, we introduce c-GNFs, how they can be used to model SCMs, and how they estimate ACEs and counterfactual inference. Next, we present the simulated experimental setup, results and analysis. 
Lastly, we conclude our work with implications in social science and beyond. 



\section{Notation, Problem Definition and Assumptions}\label{sec:problemdefinition,assumptions,notations}
In this section, we discuss the formulation of the causal problem particularly in social science, notations, and critical assumptions for causal inference.
Fig.~\ref{sfig:nh_mechanism} denotes the temporal model involving time-varying confounders used by~\citet{wodtke2011ipw} in their real-world study involving 16 waves or time-points, i.e., $k{ = }1, \ldots, 16$, where $A_k$ and $C_k$ denote the neighborhood context treatment and the observed neighborhood covariate at time $k$, and $Y$ denotes the high-school graduation as outcome of causal interest on the Panel Study of Income Dynamics (PSID) sensitive dataset~\cite{geolytics2003censuscdPSID}. 
However, in this work, we consider the $2$-wave model in Fig.~\ref{sfig:nh_mechanism 2-wave} used by~\citet{wodtke2020rwr} in his simulated experiments (non-sensitive data) such that the true causal effects are known for benchmarking of different causal effect estimation methods.

Specifically, our aim is same as that of~\citet{wodtke2020rwr}, i.e., to estimate the ACEs of the binary treatment variables $A_1$ and $A_2$ from the observational data. These ACEs ($\lambda_{a_1,a_2}$) are formally defined as 
\begin{IEEEeqnarray}{rCl}
\lambda_{1,0} = \mathbf{E}[Y_{1,0}]-\mathbf{E}[Y_{0,0}]\IEEEyesnumber\IEEEyessubnumber*
\label{eq:lambda10}\\
\lambda_{0,1} = \mathbf{E}[Y_{0,1}]-\mathbf{E}[Y_{0,0}]
\label{eq:lambda01}\\
\lambda_{1,1} = \mathbf{E}[Y_{1,1}]-\mathbf{E}[Y_{1,0}]
\label{eq:lambda11}
\end{IEEEeqnarray}
where $Y_{a_1,a_2}$ denotes the potential outcome of $Y$ under the interventions ${A_1}{ := }a_1$ and ${A_2}{ := }a_2$. Note that the previous expectations are with respect to the interventional distribution of $Y_{a_1,a_2}$ under the interventions ${A_1}{ := }a_1$ and ${A_2}{ := }a_2$
\begin{IEEEeqnarray}{rll}
P(Y_{\{a_k\}^{2}_{k=1}})\enspace.\hfill\IEEEyesnumber\IEEEyessubnumber\label{eq:wodtke2011rwr}
\end{IEEEeqnarray}
Under the assumptions we discuss next, the interventional distribution in Eq.~\eqref{eq:wodtke2011rwr} can be expressed in terms of the components from the observational joint distribution via $do$-calculus~\cite{pearl2012docalculus} or $G$-computation formula~\cite{ROBINS1986gcom, hernan2009ipw, pearl2009causality} as 
\begin{IEEEeqnarray}{rll}
P(Y_{\{a_k\}^{2}_{k=1}}) &=\sum_{{\{{C_k}\}}^{2}_{k=1}}&P(Y|{\{{C_k},{a_k}\}}^{2}_{k=1}) \IEEEnonumber*\\
 & &P(C_2|C_1,a_1) \IEEEnonumber*\\
 & &\left(\Pi_{j=1}^{2} 1_{(A_{j}{ = }a_{j})}\right) P(C_1)\enspace,\hfill\IEEEyessubnumber
\label{eq:2-wave}
\end{IEEEeqnarray}
where $1_{({A_j}{ = }a_j)}$ is the indicator function denoting the intervention ${A_j}{ := }a_j$. 
For the curious readers, the interventional distribution for the general $K$-wave model of~\citet{wodtke2011ipw} can be expressed analogously as
\begin{IEEEeqnarray}{rll}
P(Y_{\{a_k\}^{K}_{k=1}})& =\sum_{{\{{C_k}\}}^{K}_{k=1}}&P(Y|{\{{C_k},{a_k}\}}^{K}_{k=1}) \IEEEnonumber*\\
 & &\left(\Pi_{j=2}^{K} P(C_j|C_{j-1},a_{j-1})\right) \IEEEnonumber*\\
 & &\left(\Pi_{j=1}^{K} 1_{(A_{j}{ = }a_{j})}\right) P(C_1)\enspace.\hfill\IEEEyessubnumber
\label{eq:K-wave man}
\end{IEEEeqnarray}

\subsection{Critical Assumptions in Causal Inference}
Although generally `correlation does not imply causation', causal inference is about defining under what assumptions and conditions correlation coincides with causation~\cite{pearl2009causality, neal2020Intro2CI}. For our counterfactual-inference argument, we require six assumptions and conditions that are briefly stated below. 
The technical appendix shows detailing of the mathematical definitions and how causal quantities can be estimated from purely statistical quantities. 

\begin{inparaenum}[1.]
\item \textbf{Unconfoundedness} \textbf{or} \textbf{no unobserved confounders}: 
As shown in Fig.~\ref{sfig:nh_mechanism}, we assume that there is no unobserved confounding between $A_k$ and $Y$. Otherwise, it is impossible to know if the observed correlation is due to causality or confounding. This assumption is strong and untestable~\cite{Rubin1990unconfoundedness}. 

\item \textbf{Positivity} \textbf{or} \textbf{overlap} \textbf{or} \textbf{common support} \textbf{or} \textbf{extrapolation}: We assume that every individual has non-zero probability of receiving any of the treatments, i.e., $P(A_k|A_{k-1},C_k)$ is always positive. Otherwise, it is impossible to accurately estimate the causal effect as the model is inaccurate in the non-overlap region due to no data.

\item \textbf{Conditional ignorability or exchangability}: In economics and epidemiology, the assumptions of unconfoundedness and positivity are referred together as conditional ignorability or exchangability~\cite{Rosenbaum1983propensityscoreipw}. Unconditional ignorability or exchangability can only be achieved in Randomized Controlled Trials (RCTs).

\item \textbf{Consistency}: This assumption states that the observed outcome is same as the potential outcome under the observed treatment~\cite{ROBINS1986gcom, cole2009consistency, vanderweele2009concerningconsistency}.

\item \textbf{No interference} \textbf{or} \textbf{Stable Unit-Treatment Value Assumption (SUTVA)}: No interference means that a particular individual's outcome is only a function of that particular individual's treatment and is not affected by the treatment of any other individual~\cite{cox1958planning}.

\item \textbf{Modularity} \textbf{or} \textbf{independent mechanisms} \textbf{or} \textbf{autonomy} \textbf{or} \textbf{invariance}: Any joint distribution can be factorized as the product of conditional distributions corresponding to independent causal mechanisms using the Markov assumption. 
Under the interventions, the modularity assumption states that all the mechanisms of the non-intervened variables remains the same while the mechanism of the intervened variables is set to the intervened value. 
The modularity assumption can be seen in Eqs.~\eqref{eq:2-wave} and~\eqref{eq:K-wave man}, where the probability terms corresponding to the mechanism of the intervened variable ${A_j}{ := }a_j$, i.e., $P(A_j|C_{j},A_{j-1})$ term in the observational joint distribution is replaced by the indicator function $1_{({A_j}{ = }a_j)}$ denoting the intervention ${A_j}{ := }a_j$.
This modularity assumption is crucial for causal and counterfactual inference~\cite{pearl2009abductionactionprediction, pearl2009causality, pearl2018bookofwhy}.
\end{inparaenum}

\section{SCM, Encapsulated-SCM and c-GNF}\label{sec:gnf}

In this section, we present our c-GNFs and describe their connection to SCMs. An SCM consists of a set of assignment equations describing the causal relations between the random variables of a causal system, such as
\begin{IEEEeqnarray}{rll}
X_i := f_i(PA_i, U_i) \text{ with } i {=} 1,\ldots,d\label{seq:scm}\IEEEyesnumber
\end{IEEEeqnarray}
where $\{X_i\}_{i{ = }1}^d$ represents the set of observed endogenous random variables, $PA_i$ represents the set of variables that are connoting parents of $X_i$, $\{U_i\}_{i{ = }1}^d$ denotes the set of exogenous noise random variables, and $\{f_i\}_{i{ = }1}^d$ denotes the set of functions (\emph{independent mechanisms}) that generate the endogenous variable $X_i$ from its observed causes $PA_i$ and noise $U_i$. 
SCM is also referred as Functional Causal Model (FCM) due to the functional mechanisms $\{f_i\}_{i{ = }1}^d$.

Let $\mathbf{X}{ = }[X_1,\ldots,X_d]^\text{T}{ \in }\mathbb{R}^d$ denote the $d$-dimensional vector of SCM endogenous variables. Similarly, let $\mathbf{U}{ = }[U_1,\ldots,U_d]^\text{T}{ \in }\mathbb{R}^d$ denote the $d$-dimensional vector of SCM exogenous variables and $F{:}\mathbf{U}{ \rightarrow }\mathbf{X}$ denote the transformation representing the true SCM such that $\mathbf{X}{ = }F\mathbf{U}$. Let $\mathcal{G}$ represent a causal-DAG with $\{X_1,\ldots,X_d\}$ as the set of nodes and adjacency matrix $\mathcal{A_G}{ \in }\{0,1\}^{d{ \times }d}$. Let $P_\mathbf{X}(\mathbf{X})$ represent the joint distribution over the endogenous variables $\mathbf{X}$, which factorizes according to the causal-DAG $\mathcal{G}$ as
\begin{IEEEeqnarray}{rll}
P_\mathbf{X}(\mathbf{X}) = \Pi_{i=1}^{d}P({X_i}|PA_i)\enspace,\label{seq:bayesfactor}\IEEEyesnumber\IEEEyessubnumber
\label{eq:P_x joint distn bn}
\end{IEEEeqnarray}
where $PA_i{ = }\{X_j{ : }\mathcal{A_G}_{i,j}{ = }1\}$ denotes the set of parents of the vertex/node $X_i$ in the causal-DAG $\mathcal{G}$. As we show in the next section, the factorized joint distribution in Eq.~\eqref{eq:P_x joint distn bn} can be modeled using an autoregressive model parameterized by a deep-neural-network $\theta$, which we denote by $P_\mathbf{X}(\mathbf{X};\theta)$. Such a model is what we call c-GNF.

\subsection{Causal-Graphical Normalizing Flows (c-GNFs)}
Normalizing Flows (NFs)~\cite{tabak2010nf, tabak2013nf, rezende2015variationalNF, kobyzev2020NF, papamakarios2021NF_pmi} are a family of generative models with tractable distributions where both sampling and
density evaluation can be efficient and exact. 
A NF is a flow-based model with transformation $T{ : }\mathbf{X}{ \rightarrow }\mathbf{Z}$, such that $\mathbf{Z}{ = }T\mathbf{X}$, where $\mathbf{Z}{ = }[Z_1,\ldots,Z_d]^\text{T}{ \in }\mathbb{R}^d$ represents the \emph{base random variable} of the flow-model with the \emph{base distribution} $P_\mathbf{Z}(\mathbf{Z})$ that is usually a $d$-dimensional standard normal distribution for computational convenience and ease of density estimation. Hence the name \emph{normalizing flow}. The defining properties of $T$ are, (i) $T$ must be invertible with $T^{-1}$ as the inverse generative flow such that $T^{-1}{ : }\mathbf{Z}{ \rightarrow }\mathbf{X}$, and (ii) $T$ and $T^{-1}$ must be differentiable, i.e., $T$ must be a $d$-dimensional diffeomorphism~\cite{milnor1997diffeomorphism}. 
Under these properties, from the change of variables formula, we can express the endogenous joint distribution $P_\mathbf{X}(\mathbf{X})$ in terms of the assumed base distribution $P_\mathbf{Z}(\mathbf{Z})$ as
\begin{IEEEeqnarray}{rll}
P_\mathbf{X}(\mathbf{X}){ = }P_\mathbf{Z}(T\mathbf{X}) |\mathrm{det} J_{T}(\mathbf{X})|\enspace .\hfill\IEEEyessubnumber
\label{eq:p_x joint distn nf}
\end{IEEEeqnarray}
Since calculating the joint density of $P_\mathbf{X}(\mathbf{X})$ requires the calculation of the determinant of the Jacobian of $T$ with respect to $\mathbf{X}$, i.e., $\mathrm{det} J_{T}(\mathbf{X})$, it is advantageous for computational reasons to choose $T$ to have an autoregressive structure such that $J_{T}(\mathbf{X})$ is a lower-triangular matrix and $\mathrm{det} J_{T}(\mathbf{X})$ is just the product of the diagonal elements. 
Autoregressive Flows (AFs)~\cite{kingma2016IAF, Papamakarios2017MAF, Huang2018NAF} are NFs that model autoregressive structure.
AFs are composed of two components, the transformer and the conditioner~\cite{papamakarios2021NF_pmi}. Under the assumption of a strictly monotonic transformer, AFs are \emph{universal density estimators}~\cite{Huang2018NAF}. 
Of all the current AFs, Graphical Normalizing Flows (GNFs)~\cite{wehenkel2020GNF} facilitate the use a desired DAG representation as opposed to an arbitrary autoregressive structure. 
For causal inference, it is crucial that the DAG has a causal interpretation.
~In this work, we assume the true causal-DAG $\mathcal{G}$ for the GNFs, hence the term causal-GNFs (c-GNFs). 
In c-GNFs, we use Unconstrained Monotonic Neural Network (UMNN)~\cite{wehenkel2019UMNN}, a strictly monotonic integration based transformer with $\mathcal{G}$ for the graphical conditioner.

Since Markov equivalent DAGs induce equivalent factorizations of the observational joint distributions, GNFs that use Markov equivalent DAGs represent the same observational joint distribution. This is problematic for causal and counterfactual inference, as different GNFs may represent different interventional joint distributions. In other words, to perform causal inference as we do in this work, the GNF needs to use causal-DAG $\mathcal{G}$. 
Hence the name causal-GNF (c-GNF). To appreciate this, consider the Causal Autoregressive Flow (CAREFL) by~\citet{khemakhem2021causalAF}. While CAREFL uses AF, it is flawed as it considers all the predecessors in the topological ordering, i.e., a Markov equivalent DAG, when it should strictly be just the connoting parents/causes of the given node, i.e., the causal-DAG $\mathcal{G}$. 
We present an example of CAREFL's flaw in our technical appendix, which shows the conditions under which CAREFL fails in counterfactual inference. 
Note that only a GNF with the true causal-DAG $\mathcal{G}$ encapsulates the true SCM, thus satisfying the modularity assumption, which is necessary for correct causal and counterfactual inference. This is formalized as
\begin{IEEEeqnarray}{rll}
\mathbf{X}{ = }F\mathbf{U}{ = }F(H(\mathbf{Z})){ = }(FH)\mathbf{Z}{ = }\tilde{F}\mathbf{Z}{ = }T^{-1}\mathbf{Z}\enspace, \IEEEyesnumber
\label{eq:scm_cgnf}
\end{IEEEeqnarray}
where $H{ : }\mathbf{Z}{ \rightarrow }\mathbf{U}$ is an auxiliary transformation such that $U_i{ = }h_i({Z_i})$ for $i{ = }1,\ldots,d$ and $H{ = }[h_1,\ldots,h_d]^\text{T}$. %
It follows from Eq.~\eqref{eq:scm_cgnf} that our c-GNF $T^{-1}{ = }\tilde{F}{ = }(FH)$ encapsulates the true SCM $F$ as $T$ and $T^{-1}$ both encode causal-DAG $\mathcal{G}$ in the graphical conditioner, thus providing a way to indirectly model $F$ without making assumptions on $\mathbf{U}$ or the functional causal mechanisms $F$ or the auxiliary transformation $H$. This sets apart our c-GNFs from most other models for causal inference.
Fig.~\ref{sfig:c-GNF arch} shows the c-GNF architecture for causal and counterfactual inference using $T$ and $T^{-1}$.

Training a c-GNF amounts to training the deep neural networks that parameterize the transformers and conditioners. This is typically done by maximizing the $\log$-likelihood of the training dataset $\{\mathbf{X}^\ell\}^{N_s}_{\ell{ = }1}$, which is expressed as shown below, by using Eq.~\eqref{eq:p_x joint distn nf} for the summation term $P_\mathbf{X}(\mathbf{X}^\ell;\theta)$
\begin{IEEEeqnarray}{rll}
\mathcal{L(\theta)}{ = }&\sum_{\ell{ = }1}^{N_s}\mathrm{log}P_\mathbf{X}(\mathbf{X}^\ell;\theta)\IEEEyesnumber 
\label{eq:cgnf mll fn}
\end{IEEEeqnarray}
where $\theta$ denotes the parameters of the deep neural networks of the UMNN transformer and graphical conditioner, optimized using stochastic gradient descent.

\subsubsection{Gaussian Dequantization Trick}
\begin{algorithm}[tb!]
\caption{Gaussian Dequantization and Quantization}
\label{alg:dequatization quantization}
\begin{algorithmic}[1] 
\Procedure{Gaussian Dequantization}{$\{{D^\ell}\}^b_{\ell{ = }1}$}\label{alg:dequatization}
\State Generate $\tilde{D^\ell} \sim \mathcal{N}(\mu={D}^{\ell},\sigma^2=1/36)$
\State \textbf{return} Dequantized / continuous variables $\{\tilde{D^\ell}\}^b_{\ell{ = }1}$
\EndProcedure%
\Procedure{Quantization}{$\{{\tilde{D}^\ell}\}^b_{\ell{ = }1}$}\label{alg:quatization}
\State ${D}^{\ell} = \mathrm{clamp}(\mathrm{round}(\tilde{D^\ell}), \mathrm{min}{ = }0, \mathrm{max}{ = }N{ - }1)$
\State \textbf{return} Quantized / discrete variables $\{{D^\ell}\}^b_{\ell{ = }1}$
\EndProcedure%
\end{algorithmic}
\end{algorithm}
NFs naturally model continuous variables, yet  in practice, social scientists and others have to model both continuous and discrete variables, as treatments may be discrete categorical variables. 
Recently, discrete NFs have been an active field of research including techniques ranging from simple uniform dequantization, Gumbel-max dequantization to complex variational bound dequantization~\cite{Uria2013RNADETR_uniformdeq, Hoogeboom2019IDF, Tran2019IDF, ho2019flow++, ZieglerR2019VAE_dequant, Ma2019Macow, Nielsen2020DeqGap, Pawlowski2020DSCM_CI}.
Motivated by the Dirac-delta function from control theory, we propose and validate our novel Gaussian dequantization trick (see Algorithm~\ref{alg:dequatization quantization} for a discrete variable $D$ with $N$ classes/categories) to model discrete variables into NFs using the fact that NFs are strongest in modeling Gaussian distributions seamlessly. We provide the exact motivation for the Gaussian dequantization and other details in the technical appendix.

\section{Experiments, Results and Discussion}\label{sec:experiments}
\begin{figure*}[ht!]
    \begin{center}
    \includegraphics[width=\linewidth,keepaspectratio]{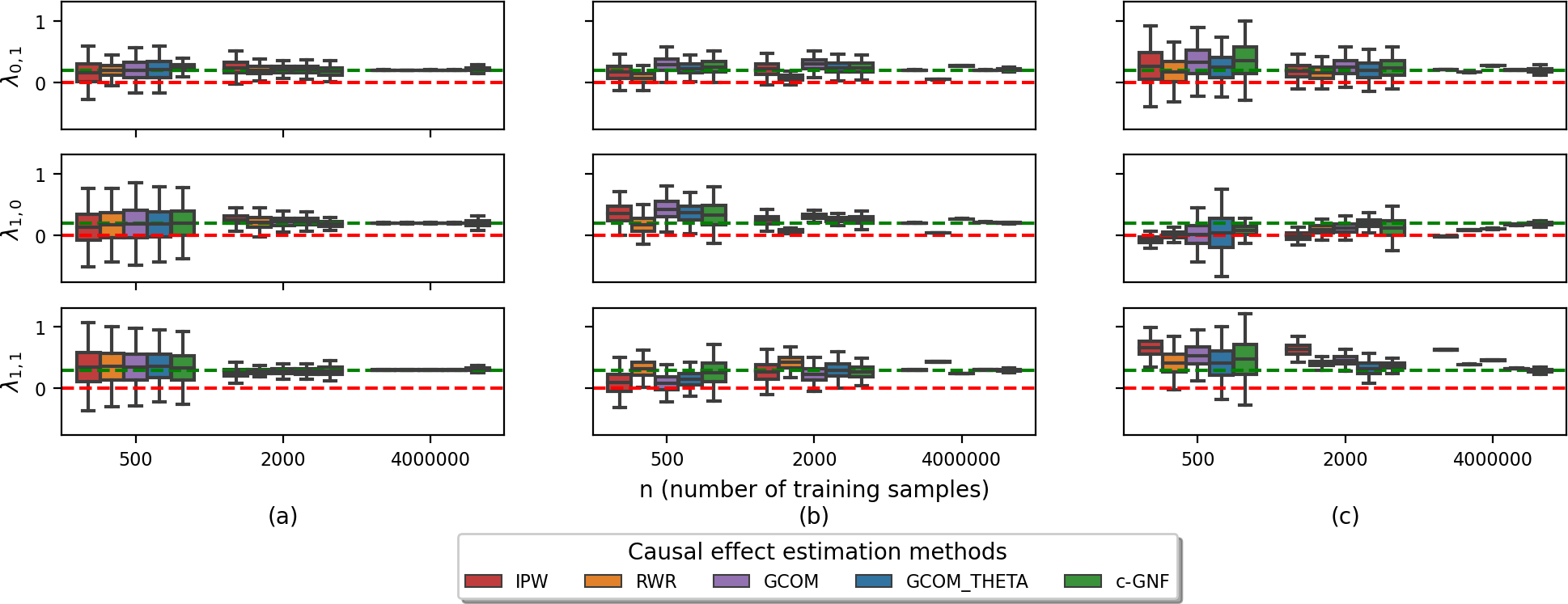}
    \caption{
(a) $\{\theta_{11}{ = }\theta_{21}{ = }0,\gamma_{21}{ = }\gamma_{12}{ = }0\}$. (b) $\{\theta_{11}{ = }\theta_{21}{ = }0.2,\gamma_{21}{ = }0.4,\gamma_{12}{ = }0\}$. (c) $\{\theta_{11}{ = }\theta_{21}{ = }0.2,\gamma_{21}{ = }\gamma_{12}{ = }0.4\}$.
}
    \label{fig:GNF main results}
    \end{center}
\end{figure*}

For our experiments and benchmarking, we simulate a virtual social system using the $2$-wave model shown in Fig.~\ref{sfig:nh_mechanism 2-wave}. We generated continuous covariates {($C_1, C_2$)}, binary treatments {($A_1, A_2$)} and a continuous outcome  $Y$.
To provide a realistic sense to the virtual social system under our study, {($C_1, C_2$)} can denote the neighborhood contexts, e.g., the parental income of the individual in the given neighborhood, and {($A_1, A_2$)} can represents neighborhood exposures, e.g., rich/poor neighborhood, and $Y$ can represent the total score obtained in high-school graduation examination.
Our aim is to analyse the neighborhood effects on the high-school graduation.
We use the following governing equations from~\citet{wodtke2020rwr} to simulate the dataset.
\begin{IEEEeqnarray}{rCl}
C_1 & \sim & \mathcal{N}(\mu{ = }0, \sigma^2{ = }1)\enspace, \IEEEyesnumber\IEEEyessubnumber*
\label{eq:c1}\\
A_1 & \sim & {Bern}(p{ = }\Phi(0.4C_1+2\gamma_{12}C^2_1))\enspace,
\label{eq:a1gc1}\\
C_2 & \sim & \mathcal{N}(\mu{ = }0.4C_1+0.2A_1, \sigma^2{ = }1)\enspace,
\label{eq:c2gc1a1}\\
A_2 & \sim & {Bern}(p{ = }\Phi(0.2A_1+0.4C_2+\gamma_{12}A^2_1 \IEEEnonumber\\
& & +2\gamma_{12}C_2+\gamma_{12}C_1A_1/2))\enspace,\IEEEyessubnumber*
\label{eq:a2gc2a1}\\
Y & \sim & \mathcal{N}(\mu{ = }0.4(C_1{ - }\mu_{C_1})+A_1(0.2+\theta_{11}C_1)\IEEEnonumber*\\
& & +(C_2{ - }\mu_{C_2})(0.4+\gamma_{21}C_1)\IEEEnonumber*\\
& & +A_2(0.2+\theta_{21}C_1+0.1A_1),\sigma^2{ = }1)\enspace,\IEEEyessubnumber*
\label{eq:ygc1a1c2a2}
\end{IEEEeqnarray}
where $\Phi$ is the standard normal cumulative distribution function, $\{\theta_{11},\theta_{21}\}$ are the parameters used to modify the magnitude of the causal effect modification/heterogeneity, and $\{\gamma_{12},\gamma_{21}\}$ are the parameter used to control the misspecification of the treatment and outcome models. 

In the work by~\citet{wodtke2020rwr}, every experimental setting is run for 10000 randomly seeded simulations with [500, 1000, 2000] training samples (typical for social science experiments) in each simulation to report the mean ($\mu_{\lambda_{a_1,a_2}}$) and standard-deviation ($\sigma_{\lambda_{a_1,a_2}}$).
However, in most practical cases, it is atypical to run deep neural networks for 10000 simulations and is limited between one to 10 in most practical deep learning applications. 
Hence, we run our experiments with different samples sizes for five randomly seeded simulations,
using all the methods (IPW, RWR, GCOM and c-GNF) and report the results pictorially in Fig.~\ref{fig:GNF main results}.
Since we run multiple simulations with multiple setting with c-GNFs, we do not concentrate on hyperparameter selection that is best for each simulation and simply use three fully-connected layers with [20, 15, 10] hidden units for the graphical conditioner and three fully-connected layers with [15, 10, 5] hidden units for the monotonic UMNN transformer.
We implement c-GNFs in Pytorch~\cite{paszke2017pytorch} using GNF baseline code\footnote{\url{https://github.com/AWehenkel/Graphical-Normalizing-Flows}} and AdamW~\cite{LoshchilovH2019adamw} optimizer with learning-rate=$1e{-3}$ and a batch-size of 128 (2GB of GPU memory) for all our experiments.
It is also our aim to evaluate the robustness of c-GNFs in estimating the ACEs without any fine hyperparameters selection using only the most basic/default settings.
Since, deep neural networks are prone to overfit, we split our data into three train, validation and test sets in ratio 8:1:1. We strictly use only the training set with [500, 2000, 4000000] samples for training and use the held-out validation set for early stopping to get the model with best validation loss. 
We further validate the generalization of the best validation loss model on the held-out test set.
We simulate 2000 samples from every interventional distributions using c-GNFs to evaluate $\mathbf{E}[Y_{a_1,a_2}]$ to obtain Monte-Carlo estimates of ACEs using Eqs.~\eqref{eq:lambda10}-\eqref{eq:lambda11}. Theoretically, a higher number of samples from the interventional distributions would provide better Monte-Carlo estimates at an added computational cost.

Fig.~\ref{fig:GNF main results} indicates the results for the following three experimental settings: 
\begin{inparaenum}[(i)]
\item Fig.~\ref{fig:GNF main results}a shows the results with both the treatment and outcome models correctly specified without heterogeneity in data.
\item Fig.~\ref{fig:GNF main results}b shows the results with only the treatment model correctly specified with heterogeneity in data.
\item Fig.~\ref{fig:GNF main results}c shows the results with both treatment and outcome models misspecified with heterogeneity in data.
\end{inparaenum}
Under all three experimental settings, the true ACEs of interest for the above $2$-wave time-varying model $\lambda_{a_1,a_2}$ for the binary treatments $\{{A_1}{ := }a_1, {A_2}{ := }a_2\}$ are calculated from Eqs.~\eqref{eq:lambda01}-\eqref{eq:lambda11} as $\lambda_{1,0}{ = }{0.2{ - }0}{ = }0.2, \lambda_{0,1}{ = }{0.2{ - }0}{ = }0.2$, and $\lambda_{1,1}{ = }{0.5{ - }0.2}{ = }0.3$ and denoted by green horizontal dotted lines in Fig.~\ref{fig:GNF main results}.
The red horizontal dotted line in Fig.~\ref{fig:GNF main results} represents \emph{zero}-ACE which can be considered as the critical point due to the fact that \emph{zero}-ACE indicates both the treatments to have the same causal effect/outcome. Sometimes, it is more important to accurately identify the sign of the ACE with some acceptable tolerance in the magnitude of the ACE estimate as the sign is what determines the effectiveness of the treatment, e.g., $\lambda_{1,0}{ > }0$ in Eq.~\eqref{eq:lambda10} indicates treatment $\{{A_1}{ := }1, {A_2}{ := }0\}$ is better than treatment $\{{A_1}{ := }0, {A_2}{ := }0\}$ in expectation, and vice-versa if $\lambda_{1,0}{ < }0$. It is of primary importance to identify which treatment is better and of secondary importance to identify by what amount the treatment is better, in expectation.
In Fig.~\ref{fig:GNF main results}, the box-plots indicate {$\mu, \mu{ \pm }\sigma,\mu{ \pm }3\sigma$} values for all $\lambda_{a_1,a_2}$ to provide an idea of the variance of the estimates.
Observe that in all the plots, c-GNF does not include the \emph{zero}-ACE in $\mu{ \pm }\sigma$ range indicating that c-GNF does not introduce ambiguity in the selection of best treatment, in expectation, unlike IPW, RWR and GCOM. This non-ambiguity is a desired quality from any causal effect estimation methods used in decision-making.

In Fig.~\ref{fig:GNF main results}a, we see that all the models perform equally well under no model misspecification and effect modification/heterogeneity in data.
However increasing the complexity by adding outcome model misspecification and data heterogeneity in Fig.~\ref{fig:GNF main results}b, we see that the regression based outcome methods not assuming the right functional forms, i.e., GCOM and RWR, result in biased estimates in both small and large samples. 
{GCOM\_THETA} in Fig.~\ref{fig:GNF main results} refers correcting GCOM by specifying the true functional form manually or parameterizing the model by a deep neural network as in the case of meta-learners to result in unbiased estimates.
Hence, we see that {GCOM\_THETA} performs better than {GCOM} due to the correction.
Similarly, since the treatment model is rightly specified, IPW estimates are unbiased. 
In contrast to Figs.~\ref{fig:GNF main results}a and \ref{fig:GNF main results}b, Fig.~\ref{fig:GNF main results}c indicates the treatment model is misspecified as well. With the treatment model misspecified, IPW no longer results in unbiased estimates.
From Fig.~\ref{fig:GNF main results}, we show that in small-sample size regime, c-GNF results in unbiased estimates with comparable variance.
We can still do better in terms of variance by appropriate selection of the deep neural networks as a part of the hyperparameter tuning, which is not our aim.
A real-world application of c-GNF on a large-scale non-randomized observational study to analyse the impact of IMF (International Monetary Fund) program on the child poverty is conducted in~\citet{balgi2022counterfactual} that observes an effective reduction of child poverty by $1.2{\pm}0.24$ degrees in the Global-South, thus indicating the beneficial nature of IMF program on child poverty reduction.

\subsection{Counterfactual Inference}
\begin{figure}[ht!]
\centering
\begin{subfigure}[t]{0.45\textwidth}
\includegraphics[width=\linewidth,height=\linewidth,keepaspectratio]{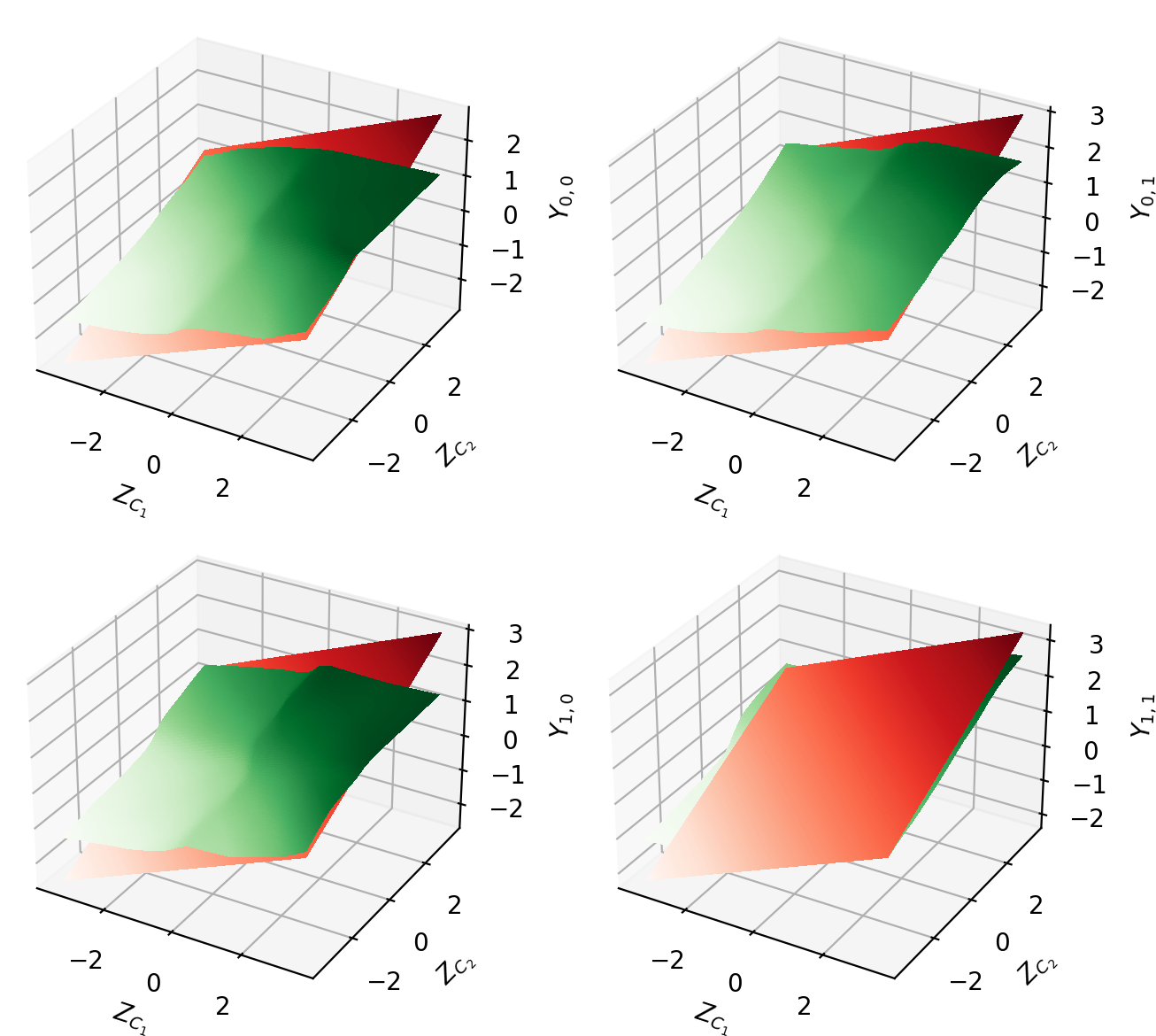}
\caption{}
    \label{sfig:potential outcomes map}
\end{subfigure}
\hfill
\centering
\begin{subfigure}[t]{0.23\textwidth}
\includegraphics[width=\linewidth,height=\linewidth,keepaspectratio]{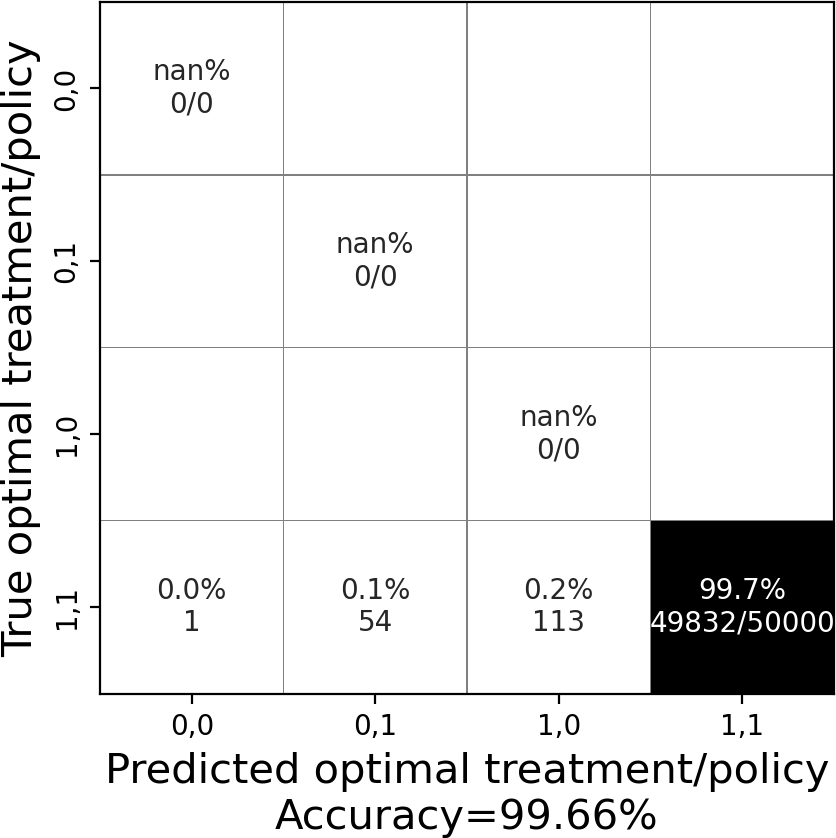}
    \caption{}
    \label{sfig:personalized optimal treatment val}
\end{subfigure}
\hfill
\centering
\begin{subfigure}[t]{0.23\textwidth}
\includegraphics[width=\linewidth,height=\linewidth,keepaspectratio]{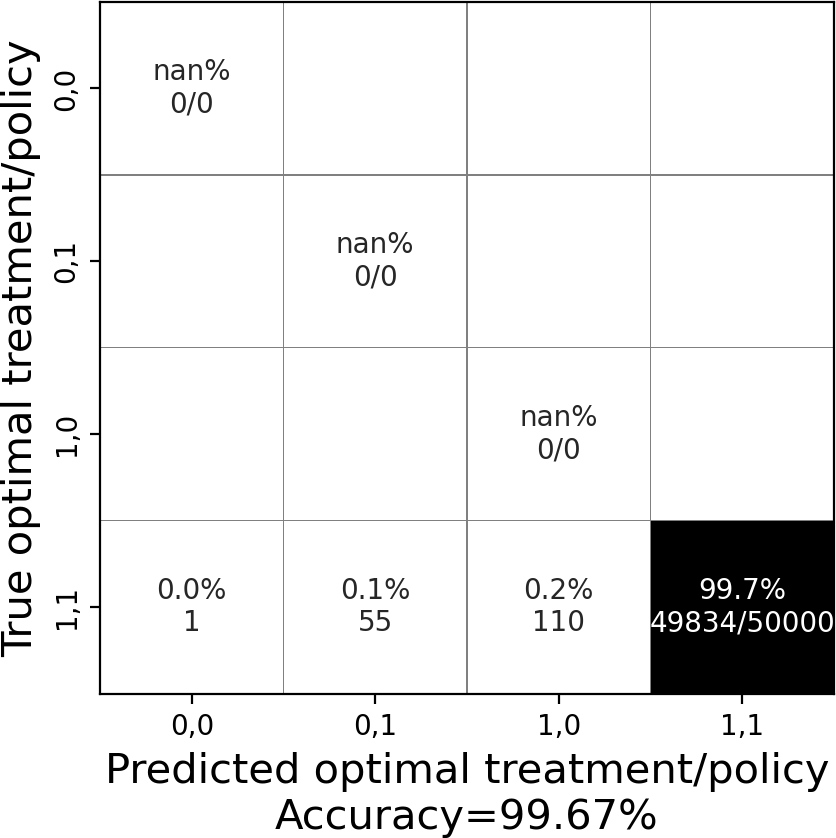}
    \caption{}
    \label{sfig:personalized optimal treatment tst}
\end{subfigure}
\hfill
    \caption{
    (a) The red and green surface-plots respectively represent the potential outcome maps of true SCM $F$ and our \emph{encapsulated-SCM} $\tilde{F}$ generated from the SCM noise ranges $Z_{C_1}{ \in }[-3,3], Z_{C_2}{ \in }[-3,3], Z_Y{ = }0$.
    ~(b, c) Confusion matrices between the true and predicted optimal personalised policy on unseen validation and test sets respectively (with model under no misspecification setting from Fig.~\ref{fig:GNF main results}a).
    }
    \label{fig:first law of causal inference}
\end{figure}
Apart from the benefits of ACE estimates with small bias and variance compared to standard statistical methods, c-GNF provides the unique benefit of counterfactual inference using \emph{`The First Law of Causal Inference'}~\cite{pearl2009abductionactionprediction, pearl2009causality, pearl2018bookofwhy}, which is virtually non-existent in IPW/RWR/GCOM. `The First Law of Causal Inference' essentially provides a framework to identify the unit level potential outcomes using SCM, thereby addressing the fundamental missing value problem of counterfactual inference. This law to identify the missing/unseen potential outcome $Y_{a_1,a_2}$ for a unit $\mathbf{Z^\ell}$ is represented mathematically as
\begin{IEEEeqnarray}{rll}
Y_{a_1,a_2}(\mathbf{Z}^\ell){ = }Y_{\tilde{F}_{a_1,a_2}}(\mathbf{Z}^\ell)\enspace,\IEEEyesnumber\label{eq:first law of causal inference}
\end{IEEEeqnarray}
where $Y_{a_1,a_2}(\mathbf{Z}^\ell)$ denotes the potential outcome $Y$ under the treatments $\{{A_j}{ := }a_j\}^2_{j{ = }1}$ for a given unit individual $\mathbf{Z}^\ell$ that is fundamentally missing and is of interest for counterfactual inference, and $Y_{\tilde{F}_{a_1,a_2}}(\mathbf{Z}^\ell)$ represents the actual outcome $Y$ observed from the mutilated-SCM $\tilde{F}_{a_1,a_2}$ that is obtained by the mutilation $\{{A_j}{ := }a_j\}^2_{j{ = }1}$ on the \emph{encapsulated-SCM} $\tilde{F}$. Essentially, this law involves three steps that help identify the missing potential outcomes, thereby addressing the fundamental missing value problem of causal inference.
\begin{inparaenum}

    \item \textbf{Abduction}: For a given observed unit-individual evidence $\mathbf{X}^\ell$, the respective exogenous SCM noise $\mathbf{Z}^\ell$ corresponding to $\mathbf{X}^\ell$ is recovered from the encapsulated-SCM $\tilde{F}$. In case of c-GNF, we have $\mathbf{Z}^\ell{ = }T\mathbf{X}^\ell{ = }\tilde{F}^{-1}\mathbf{X}^\ell$ where the \emph{encapsulated-SCM} noise $\mathbf{Z}^\ell$ 
   indicates the unique hidden essence/identifier/DNA
    of the $\ell^{th}$ unit-individual with observed evidence $\mathbf{X}^\ell$.
    
    \item \textbf{Action}: The action or intervention (in our case a public policy) corresponding to the desired treatment is conducted in the form of mutilating the corresponding structural equation of the treatment $\{{A_j}{ := }a_j\}^2_{j{ = }1}$ in $\tilde{F}$ (modularity assumption) resulting in the mutilated-SCM $\tilde{F}_{a_1,a_2}$. 
    
    \item \textbf{Prediction}: The recovered noises $\mathbf{Z}^\ell$ from abduction are re-propagated through $\tilde{F}_{a_1,a_2}$ and the outcomes $Y_{\tilde{F}_{a_1,a_2}}(\mathbf{Z}^\ell)$ observed are nothing but the potential outcomes $Y_{a_1,a_2}(\mathbf{Z}^\ell)$ we were interested to begin with.
\end{inparaenum}    


The potential outcome maps in Fig.~\ref{sfig:potential  outcomes map} are generated using the three steps of counterfactual inference with c-GNF, i.e., \emph{encapsulated-SCM} and the noises $Z_{C_1}{ \in }[-3,3], Z_{C_2}{ \in }[-3,3], Z_Y{ = }0$ to observe the effect of $C_1$ and $C_2$ on $Y$.
Due to the linear nature of the structural equation in Eq.~\eqref{eq:ygc1a1c2a2} under no model misspecification setting, the true potential outcome maps are represented by the flat hyperplanes in Fig.~\ref{sfig:potential outcomes map} (red).
Observe that around the corners of the green 3D surface-plots which represents the low likelihood region of observing samples, the extrapolation assumption is slightly violated as the observational data corresponding to these regions are likely missing in the training set due to low likelihood. 
Extrapolation is an inevitable assumption made in almost all statistical models which is almost always violated under practical finite data setting.
True extrapolation can be attained only with infinite data, which is impractical and unrealistic.
~In the technical appendix, we further present the same set of plots as in Fig.~\ref{fig:first law of causal inference} under the complex experimental setting when both the models are misspecified and also the data exhibits effect modification/heterogeneity.

The confusion matrices from unseen validation and test sets in Figs.~\ref{sfig:personalized optimal treatment val} and~\ref{sfig:personalized optimal treatment tst} indicate the true and predicted optimal individual-level treatment/policy $\bar{a}^\ell_1,\bar{a}^\ell_2$ for the $\ell^{th}$ individual $\mathbf{X}^\ell$, which is predicted as below from Eq.~\eqref{eq:first law of causal inference}
\begin{IEEEeqnarray}{rll}
\bar{a}^\ell_1,\bar{a}^\ell_2{ = }\argmax_{a_1,a_2} Y_{a_1,a_2}(T\mathbf{X}^\ell)\enspace.\IEEEyesnumber\label{eq:optimal personalized treatment}
\end{IEEEeqnarray}
Using c-GNF, we observe that the optimal treatments are indeed predicted with a very high accuracy of 99.6\% on unseen validation and test sets under no-misspecification and no effect modification/heterogeneity in data. 
Similarly, under both model misspecifications we obtain 91\% accuracy even with our sub-optimal hyperparameters.
Thus, c-GNF enables precise P$^3$A targeting and articulation of the tailored treatments/policies in the social sciences, instead of relying on `one-size-fits-all' sub-optimal treatments/policies.

Finally, our c-GNF not only model the encapsulation of the true SCM, but also offer exact density estimation which provides necessary tools for causal inference practitioners to identify other causal quantities such as 
the probabilities of causation~\cite{pearl1999probabilitiesofcausation, pearl2009causality}, i.e., Probability of Necessity (PN), Probability of Sufficiency (PS), Probability of Necessity and Sufficiency (PNS), Probability of Disablement (PD), Probability of Enablement (PE), 
that are of utmost importance to policy-makers planning various social programs.

\section{Conclusion}\label{sec:conclusion}

This article developed causal-Graphical Normalizing Flow (\emph{c-GNF}) for personalized public policy analysis (P$^3$A). We demonstrated that our c-GNF learnt using only observational data without any assumptions on the exogenous noises or the functional mechanisms of the underlying SCM.
We further identified c-GNF is on/above-par well-established standard causal effect estimation methods such as IPW and RWR.
Most importantly, in contrast to IPW and RWR, using simulation experiments, we demonstrated the unique benefit of counterfactual inference using \emph{`The First Law of Causal Inference'} with c-GNF. Although our simulation study might be perceived as a limitation, the strong benchmarking of c-GNF on the simulated dataset demonstrated the usability of our framework with real-world data. 
The counterfactual inference demonstrated with c-GNFs showed potential to identify the optimal treatments at an individual-level, enabling precise P$^3$A. When successful, P$^3$A will likely accelerate the use of individualized policies beyond the closed settings of medical applications and into the open settings of the social science.

\section*{Acknowledgments}
The authors would like to acknowledge the generous funding from the Swedish Research Council through the Swedish Network for Register-Based Research (SWE-REG) with respect to the grant number 2019-00245 with an aim to advance machine learning to tackle problems in social sciences through register-based-learning using the Swedish population registry.

\section*{Ethics Statement}
The authors would like to indicate that the current work does not involve any real ethical considerations as assumptions are clearly stated and the experimental results with benchmarking are based only on the non-sensitive simulated dataset.

\bibliography{wodtke_2018}

\begin{thebibliography}{50}
\providecommand{\natexlab}[1]{#1}

\bibitem[{Balgi, Pe{\~n}a, and Daoud(2022)}]{balgi2022counterfactual}
Balgi, S.; Pe{\~n}a, J.~M.; and Daoud, A. 2022.
\newblock Counterfactual Analysis of the Impact of the IMF Program on Child
  Poverty in the Global-South Region using Causal-Graphical Normalizing Flows.
\newblock \emph{arXiv preprint arXiv:2202.09391}.

\bibitem[{Bang and Robins(2005)}]{bang2005doublyrobust}
Bang, H.; and Robins, J.~M. 2005.
\newblock Doubly Robust Estimation in Missing Data and Causal Inference Models.
\newblock \emph{Biometrics}, 61(4): 962--972.

\bibitem[{Cole and Frangakis(2009)}]{cole2009consistency}
Cole, S.~R.; and Frangakis, C.~E. 2009.
\newblock The Consistency Statement in Causal Inference: A Definition or An
  Assumption?
\newblock \emph{Epidemiology}, 20(1): 3--5.

\bibitem[{Cox(1958)}]{cox1958planning}
Cox, D.~R. 1958.
\newblock \emph{Planning of Experiments}.
\newblock New York: Wiley.

\bibitem[{Fienberg and Duncan(1975)}]{Fienberg1975Introduction2SEM}
Fienberg, S.; and Duncan, O.~D. 1975.
\newblock Introduction to Structural Equation Models.
\newblock \emph{Journal of the American Statistical Association (JASA)}, 72:
  485.

\bibitem[{Fisher(1936)}]{fisher1936designofexperiments}
Fisher, R.~A. 1936.
\newblock Design of Experiments.
\newblock \emph{British Medical Journal (BMJ)}, 1(3923): 554--554.

\bibitem[{GeoLytics(2003)}]{geolytics2003censuscdPSID}
GeoLytics, I. 2003.
\newblock CensusCD Neighborhood Change Database, 1970--2000 Tract Data.
\newblock \emph{New Brunswick, NJ: GeoLytics}.

\bibitem[{Goldberger(1972)}]{Goldberger1972SEMecon}
Goldberger, A.~S. 1972.
\newblock Structural Equation Methods in the Social Sciences.
\newblock \emph{Econometrica}, 40(6): 979--1001.

\bibitem[{Haavelmo(1943)}]{Haavelmo1943SEM}
Haavelmo, T. 1943.
\newblock The Statistical Implications of a System of Simultaneous Equations.
\newblock \emph{Econometrica}, 11: 1--12.

\bibitem[{Hernán and Robins(2009)}]{hernan2009ipw}
Hernán, M.~A.; and Robins, J.~M. 2009.
\newblock \emph{Causal Inference: What If}.
\newblock Boca Raton: Chapman \& Hall/CRC.

\bibitem[{Ho et~al.(2019)Ho, Chen, Srinivas, Duan, and Abbeel}]{ho2019flow++}
Ho, J.; Chen, X.; Srinivas, A.; Duan, Y.; and Abbeel, P. 2019.
\newblock Flow++: Improving Flow-Based Generative Models with Variational
  Dequantization and Architecture Design.
\newblock In \emph{International Conference on Machine Learning (ICML)},
  2722--2730.

\bibitem[{Hoogeboom et~al.(2019)Hoogeboom, Peters, van~den Berg, and
  Welling}]{Hoogeboom2019IDF}
Hoogeboom, E.; Peters, J. W.~T.; van~den Berg, R.; and Welling, M. 2019.
\newblock Integer Discrete Flows and Lossless Compression.
\newblock In \emph{Neural Information Processing Systems (NeurIPS)},
  12134--12144.

\bibitem[{Huang et~al.(2018)Huang, Krueger, Lacoste, and
  Courville}]{Huang2018NAF}
Huang, C.; Krueger, D.; Lacoste, A.; and Courville, A.~C. 2018.
\newblock Neural Autoregressive Flows.
\newblock In \emph{International Conference on Machine Learning (ICML)},
  2083--2092.

\bibitem[{Khemakhem et~al.(2021)Khemakhem, Monti, Leech, and
  Hyvarinen}]{khemakhem2021causalAF}
Khemakhem, I.; Monti, R.; Leech, R.; and Hyvarinen, A. 2021.
\newblock Causal Autoregressive Flows.
\newblock In \emph{International Conference on Artificial Intelligence and
  Statistics (AISTATS)}, 3520--3528.

\bibitem[{King(1974)}]{Duncan1974SEMsocio}
King, M. 1974.
\newblock {A. S. Goldberger and O. D. Duncan. Structural Equation Models in the
  Social Sciences}.
\newblock \emph{The Economic Journal}, 84(333): 212--214.

\bibitem[{Kingma et~al.(2016)Kingma, Salimans, Jozefowicz, Chen, Sutskever, and
  Welling}]{kingma2016IAF}
Kingma, D.~P.; Salimans, T.; Jozefowicz, R.; Chen, X.; Sutskever, I.; and
  Welling, M. 2016.
\newblock Improved Variational Inference with Inverse Autoregressive Flow.
\newblock In \emph{Neural Information Processing Systems (NeurIPS)},
  4743--4751.

\bibitem[{Kino et~al.(2021)Kino, Hsu, Shiba, Chien, Mita, Kawachi, and
  Daoud}]{kino2021adelmlreview}
Kino, S.; Hsu, Y.-T.; Shiba, K.; Chien, Y.-S.; Mita, C.; Kawachi, I.; and
  Daoud, A. 2021.
\newblock A Scoping Review on the Use of Machine Learning in Research on Social
  Determinants of Health: Trends and Research Prospects.
\newblock \emph{Social Science \& Medicine - Population Health (SSM-PH)}, 15:
  100836--100855.

\bibitem[{Kobyzev, Prince, and Brubaker(2020)}]{kobyzev2020NF}
Kobyzev, I.; Prince, S.; and Brubaker, M. 2020.
\newblock Normalizing Flows: An Introduction and Review of Current Methods.
\newblock \emph{IEEE Transactions on Pattern Analysis and Machine Intelligence
  (TPAMI)}, 1--1.

\bibitem[{K{\"u}nzel et~al.(2019)K{\"u}nzel, Sekhon, Bickel, and
  Yu}]{kunzel2019metalearners}
K{\"u}nzel, S.~R.; Sekhon, J.~S.; Bickel, P.~J.; and Yu, B. 2019.
\newblock Metalearners for Estimating Heterogeneous Treatment Effects using
  Machine Learning.
\newblock \emph{Proceedings of the National Academy of Sciences (PNAS)},
  116(10): 4156--4165.

\bibitem[{Loshchilov and Hutter(2019)}]{LoshchilovH2019adamw}
Loshchilov, I.; and Hutter, F. 2019.
\newblock Decoupled Weight Decay Regularization.
\newblock In \emph{International Conference on Learning Representations
  (ICLR)}.

\bibitem[{Ma et~al.(2019)Ma, Kong, Zhang, and Hovy}]{Ma2019Macow}
Ma, X.; Kong, X.; Zhang, S.; and Hovy, E.~H. 2019.
\newblock MaCow: Masked Convolutional Generative Flow.
\newblock In \emph{Neural Information Processing Systems (NeurIPS)},
  5891--5900.

\bibitem[{Milnor and Weaver(1997)}]{milnor1997diffeomorphism}
Milnor, J.; and Weaver, D.~W. 1997.
\newblock \emph{Topology from the Differentiable Viewpoint}.
\newblock Princeton university press.

\bibitem[{Neal(2020)}]{neal2020Intro2CI}
Neal, B. 2020.
\newblock Introduction to Causal Inference from a Machine Learning Perspective.
\newblock \emph{Course Lecture Notes (draft)}.

\bibitem[{Nielsen and Winther(2020)}]{Nielsen2020DeqGap}
Nielsen, D.; and Winther, O. 2020.
\newblock Closing the Dequantization Gap: PixelCNN as a Single-Layer Flow.
\newblock In \emph{Neural Information Processing Systems (NeurIPS)}.

\bibitem[{Papamakarios, Murray, and Pavlakou(2017)}]{Papamakarios2017MAF}
Papamakarios, G.; Murray, I.; and Pavlakou, T. 2017.
\newblock Masked Autoregressive Flow for Density Estimation.
\newblock In \emph{Neural Information Processing Systems (NeurIPS)},
  2338--2347.

\bibitem[{Papamakarios et~al.(2021)Papamakarios, Nalisnick, Rezende, Mohamed,
  and Lakshminarayanan}]{papamakarios2021NF_pmi}
Papamakarios, G.; Nalisnick, E.; Rezende, D.~J.; Mohamed, S.; and
  Lakshminarayanan, B. 2021.
\newblock Normalizing Flows for Probabilistic Modeling and Inference.
\newblock \emph{Journal of Machine Learning Research (JMLR)}, 22(57): 1--64.

\bibitem[{Paszke et~al.(2017)Paszke, Gross, Chintala, Chanan, Yang, DeVito,
  Lin, Desmaison, Antiga, and Lerer}]{paszke2017pytorch}
Paszke, A.; Gross, S.; Chintala, S.; Chanan, G.; Yang, E.; DeVito, Z.; Lin, Z.;
  Desmaison, A.; Antiga, L.; and Lerer, A. 2017.
\newblock {Automatic Differentiation in PyTorch}.
\newblock \emph{NeurIPS Workshops}.

\bibitem[{Pawlowski, de~Castro, and Glocker(2020)}]{Pawlowski2020DSCM_CI}
Pawlowski, N.; de~Castro, D.~C.; and Glocker, B. 2020.
\newblock Deep Structural Causal Models for Tractable Counterfactual Inference.
\newblock In \emph{Neural Information Processing Systems (NeurIPS)}.

\bibitem[{Pearl(1999)}]{pearl1999probabilitiesofcausation}
Pearl, J. 1999.
\newblock Probabilities of Causation: Three Counterfactual Interpretations and
  Their Identification.
\newblock \emph{Synthese}, 121(1): 93--149.

\bibitem[{Pearl(2009{\natexlab{a}})}]{pearl2009abductionactionprediction}
Pearl, J. 2009{\natexlab{a}}.
\newblock Causal Inference in Statistics: An Overview.
\newblock \emph{Statistics Surveys}, 3: 96--146.

\bibitem[{Pearl(2009{\natexlab{b}})}]{pearl2009causality}
Pearl, J. 2009{\natexlab{b}}.
\newblock \emph{Causality: Models, Reasoning and Inference}.
\newblock USA: Cambridge University Press, 2nd edition.
\newblock ISBN 052189560X.

\bibitem[{Pearl(2012)}]{pearl2012docalculus}
Pearl, J. 2012.
\newblock The $Do$-Calculus Revisited.
\newblock In \emph{Uncertainty in Artificial Intelligence (UAI)}, 3–11.

\bibitem[{Pearl and Mackenzie(2018)}]{pearl2018bookofwhy}
Pearl, J.; and Mackenzie, D. 2018.
\newblock \emph{The Book of Why: The New Science of Cause and Effect}.
\newblock USA: Basic Books, Inc., 1st edition.
\newblock ISBN 046509760X.

\bibitem[{Ploch, Goldberger, and Duncan(1975)}]{Ploch1975SEM}
Ploch, D.~R.; Goldberger, A.~S.; and Duncan, O.~D. 1975.
\newblock Structural Equations Models in the Social Sciences.
\newblock \emph{Social Forces}, 54: 503.

\bibitem[{Rezende and Mohamed(2015)}]{rezende2015variationalNF}
Rezende, D.; and Mohamed, S. 2015.
\newblock Variational Inference with Normalizing Flows.
\newblock In \emph{International Conference on Machine Learning (ICML)},
  1530--1538.

\bibitem[{Robins(1986)}]{ROBINS1986gcom}
Robins, J.~M. 1986.
\newblock A New Approach to Causal Inference in Mortality Studies with a
  Sustained Exposure Period—Application to Control of the Healthy Worker
  Survivor Effect.
\newblock \emph{Mathematical Modelling}, 7(9): 1393--1512.

\bibitem[{Rosenbaum and Rubin(1983)}]{Rosenbaum1983propensityscoreipw}
Rosenbaum, P.~R.; and Rubin, D.~B. 1983.
\newblock The Central Role of the Propensity Score in Observational Studies for
  Causal Effects.
\newblock \emph{Biometrika}, 70(1): 41--55.

\bibitem[{Rubin(1990)}]{Rubin1990unconfoundedness}
Rubin, D.~B. 1990.
\newblock Formal Mode of Statistical Inference for Causal Effects.
\newblock \emph{Journal of Statistical Planning and Inference (JSPI)}, 25(3):
  279--292.

\bibitem[{Tabak and Turner(2013)}]{tabak2013nf}
Tabak, E.; and Turner, C. 2013.
\newblock A Family of Nonparametric Density Estimation Algorithms.
\newblock \emph{Communications on Pure and Applied Mathematics}, 66(2):
  145--164.

\bibitem[{Tabak and Vanden-Eijnden(2010)}]{tabak2010nf}
Tabak, E.; and Vanden-Eijnden, E. 2010.
\newblock Density Estimation by Dual Ascent of the Log-Likelihood.
\newblock \emph{Communications in Mathematical Sciences}, 8(1): 217--233.

\bibitem[{Tran et~al.(2019)Tran, Vafa, Agrawal, Dinh, and Poole}]{Tran2019IDF}
Tran, D.; Vafa, K.; Agrawal, K.~K.; Dinh, L.; and Poole, B. 2019.
\newblock Discrete Flows: Invertible Generative Models of Discrete Data.
\newblock In \emph{Neural Information Processing Systems (NeurIPS)},
  14692--14701.

\bibitem[{Uria, Murray, and Larochelle(2013)}]{Uria2013RNADETR_uniformdeq}
Uria, B.; Murray, I.; and Larochelle, H. 2013.
\newblock {RNADE:} The Real-Valued Neural Autoregressive Density-Estimator.
\newblock In \emph{Neural Information Processing Systems (NeurIPS)},
  2175--2183.

\bibitem[{VanderWeele(2009{\natexlab{a}})}]{vanderweele2009concerningconsistency}
VanderWeele, T.~J. 2009{\natexlab{a}}.
\newblock Concerning the Consistency Assumption in Causal Inference.
\newblock \emph{Epidemiology}, 20(6): 880--883.

\bibitem[{VanderWeele(2009{\natexlab{b}})}]{vanderweele2009distinctionheterogeneity}
VanderWeele, T.~J. 2009{\natexlab{b}}.
\newblock On the Distinction Between Interaction and Effect Modification.
\newblock \emph{Epidemiology}, 20(6): 863--871.

\bibitem[{Wehenkel and Louppe(2019)}]{wehenkel2019UMNN}
Wehenkel, A.; and Louppe, G. 2019.
\newblock Unconstrained Monotonic Neural Networks.
\newblock In \emph{Neural Information Processing Systems (NeurIPS)},
  1545--1555.

\bibitem[{Wehenkel and Louppe(2021)}]{wehenkel2020GNF}
Wehenkel, A.; and Louppe, G. 2021.
\newblock Graphical Normalizing Flows.
\newblock In \emph{International Conference on Artificial Intelligence and
  Statistics (AISTATS)}, 37--45.

\bibitem[{Wodtke(2020)}]{wodtke2020rwr}
Wodtke, G.~T. 2020.
\newblock Regression-Based Adjustment for Time-Varying Confounders.
\newblock \emph{Sociological Methods \& Research (SMR)}, 49(4): 906--946.

\bibitem[{Wodtke, Harding, and Elwert(2011)}]{wodtke2011ipw}
Wodtke, G.~T.; Harding, D.~J.; and Elwert, F. 2011.
\newblock Neighborhood Effects in Temporal Perspective: The Impact of Long-Term
  Exposure to Concentrated Disadvantage on High School Graduation.
\newblock \emph{American Sociological Review (ASR)}, 76(5): 713--736.
\newblock PMID: 22879678.

\bibitem[{Wright(1921)}]{wright1921correlationcausation}
Wright, S. 1921.
\newblock Correlation and Causation.
\newblock \emph{Journal of Agricultural Research (JAR)}, 20: 557--585.

\bibitem[{Ziegler and Rush(2019)}]{ZieglerR2019VAE_dequant}
Ziegler, Z.~M.; and Rush, A.~M. 2019.
\newblock {Latent Normalizing Flows for Discrete Sequences}.
\newblock In \emph{{International Conference on Machine Learning (ICML)}},
  7673--7682.

\end{thebibliography}

\end{document}